\documentclass[conference, onecolumn]{IEEEtran}
\newcommand{\vtheta}{\ensuremath{\mathrm{\boldsymbol{\theta}}}}

\newcommand{\vx}{\ensuremath{\mathrm{\bold{x}}}}
\newcommand{\vy}{\ensuremath{\mathrm{\bold{y}}}}

\usepackage{stackengine}

\usepackage{cite}
\usepackage{amsmath,amssymb,amsfonts}
\usepackage{algorithmic}
\usepackage{graphicx}
\usepackage{textcomp}
\usepackage{xcolor}
\usepackage{svg}
\usepackage{setspace}
\usepackage[inline]{enumitem}
\usepackage[font=footnotesize]{caption}
\usepackage{subcaption}
\def\BibTeX{{\rm B\kern-.05em{\sc i\kern-.025em b}\kern-.08em
    T\kern-.1667em\lower.7ex\hbox{E}\kern-.125emX}}

\begin{document}

\bstctlcite{IEEEexample:BSTcontrol}
\title{\huge
Enhancing Reliability of Neural Networks at the Edge: Inverted Normalization with Stochastic Affine Transformations
}

\author{
\IEEEauthorblockN{Soyed Tuhin Ahmed\IEEEauthorrefmark{9}\IEEEauthorrefmark{2}, Kamal Danouchi\IEEEauthorrefmark{3}, 
Guillaume Prenat\IEEEauthorrefmark{3}, Lorena Anghel\IEEEauthorrefmark{3}, Mehdi B. Tahoori\IEEEauthorrefmark{2}\\}
\IEEEauthorblockA{\IEEEauthorrefmark{2}Karlsruhe Institute of Technology, Karlsruhe, Germany, \IEEEauthorrefmark{9}corresponding author, email: soyed.ahmed@kit.edu}
\IEEEauthorblockA{\IEEEauthorrefmark{3}Univ. Grenoble Alpes, CEA, CNRS, Grenoble INP, and IRIG-Spintec, Grenoble, France}
\vspace{-1em}
}

\maketitle

\begin{abstract}


Bayesian Neural Networks (BayNNs) naturally provide uncertainty in their predictions, making them a suitable choice in safety-critical applications. 
Additionally, their realization using memristor-based in-memory computing (IMC) architectures enables them for resource-constrained edge applications.
In addition to predictive uncertainty, however, the ability to be inherently robust to noise in computation is also essential to ensure functional safety. In particular, memristor-based IMCs are susceptible to various sources of non-idealities such as manufacturing and runtime variations, drift, and failure,
which can significantly reduce inference accuracy. 
In this paper, we propose a method to inherently enhance the robustness and inference accuracy of BayNNs deployed in IMC architectures.
To achieve this, we introduce a novel normalization layer combined with stochastic affine transformations.  Empirical results in various benchmark datasets show a graceful degradation in inference accuracy, with an improvement of up to $58.11\%$.
\end{abstract}


\section{Introduction}

Bayesian neural networks (BayNNs) are a type of neural network that incorporates probabilistic modeling, allowing them to estimate uncertainty in predictions and provide more robust and informative results.
Uncertainty estimation is vital in improving reliability and confidence in predictions, particularly in safety-critical applications such as autonomous driving and medical diagnosis. However, this advantage comes at the cost of higher overhead in terms of power consumption, memory usage, and latency, depending on the approximation method used.

BayNNs are a natural fit for in-memory computing (IMC) architectures with resistive non-volatile memories (NVMs) as they offer to reduce some of their inherent costs. In the IMC architecture, the matrix-vector multiplication operation of a neural network (NN) can be carried out where the data already reside. Therefore, alleviating the data movement for those operations and reducing associated energy costs. Additionally, one can leverage NVM non-idealities such as resistance variation and stochastic switching for efficient BayNN inference computations \cite{soyed_nanoarch22}. 

Furthermore, NVMs such as spin transfer torque magnetic random access memories (STT-MRAM) offer benefits such as writing speed (within a few nanoseconds), exceptional endurance, and low power consumption. In addition, it demonstrates compatibility with established semiconductor manufacturing processes~\cite{lee_world-most_2022}.

Although NVMs offer several advantages for BayNN implementations, mitigating the reliability challenges associated with these devices is of utmost importance. 
Non-idealities such as manufacturing and runtime variations, defects, and failures can significantly reduce the accuracy of inference~\cite{tsai2020robust, ye2023improving}. In safety-critical applications, it is essential not only to give uncertainty in prediction but also to maintain high accuracy even in the presence of these non-idealities.

The primary focus of existing IMC-implemented Bayesian NNs~\cite{faria_implementing_2018, ahmed2023spinbayes, ahmed2023spatial, spindrop, soyed_DATE23} has largely been on uncertainty estimation rather than improving \emph{inherent fault tolerance} against non-idealities in IMC architectures. Several other studies~\cite{9586160, 9586115, ye2023improving} utilize the Bayesian framework, exploit the NVM device variation model during training to achieve resilience to device variation, or perform neural architecture search (NAS) to find a robust network.
However, the uncertainty estimation aspect is overlooked.
Furthermore, a separate NAS may potentially be a necessity in the case of other non-idealities or NVM technology.
This, in turn, can result in a significantly computationally intensive approach.

In this paper, our aim is to close this gap by offering \emph{inherently self-immune} BayNN that 
\begin{enumerate*}
    \item does not require any implicit non-idealities modeling,
    \item is generalizable across different NVM technologies and non-idealities 
    \item easier to train,
    \item IMC implementation friendly, and
    \item is still able to provide uncertainty estimates without reducing its accuracy
\end{enumerate*}. These properties are of critical importance to ensure the reliable deployment of BayNNs implemented with IMC in safety-critical applications.

To achieve this, we introduce an \emph{inverted normalization layer} combined with stochastic affine transformations. In conventional normalization layers, affine transformations, multiplication of weight followed by addition of bias, are carried out after normalization. In our approach, this operation is reversed, and we randomly drop weights and biases of the normalization layer. This not only enhances the model's ability to provide accurate inferences but also makes it inherently more robust to the non-ideal conditions associated with IMC architectures and NVM technologies. We perform a thorough evaluation of our approach on various deep learning tasks, device non-ideality models, and NN models to show the effectiveness of our approach. Compared to existing Dropout-based IMC~\cite{ahmed2023spatial, spindrop}, we were able to improve predictive performance by up to $55.62\%$ without compromising uncertainty estimation capabilities.

The remainder of this paper is structured as follows. Section~\ref{sec:background} background and discusses related work; Section~\ref{sec:methodology} presents the methodology; Section~\ref{sec:results} provides experimental results; and finally, Section~\ref{sec:conclusion} concludes the paper and discusses future work.

\section{Background}\label{sec:background}

\subsection{Bayesian Neural Networks}
Unlike conventional Neural Networks (NNs), BayNNs offer a principled way to capture the uncertainty of the model, thereby enabling more reliable decision-making. Employing Dropout as an approximation for the BayNNs has been shown to reduce the memory requirement to that of NNs. Nevertheless, deploying BayNNs on resource-constrained devices, edge applications, and in scenarios demanding high throughput is challenging.

Bayesian neural networks are probabilistic deep learning models, in contrast to classical NNs, which represent their parameters by real-valued numbers (point estimates), Bayesian NNs consider a probability distribution over their parameters.
Consequently, Bayesian NNs consider the parameter vector $\vtheta$ as a random variable with a distribution $p(\vtheta)$. In the Bayesian framework, learning is essentially the process of estimating the posterior distribution $\vtheta \sim p(\vtheta \mid \mathcal{D})$, which represents our updated beliefs about the model parameters $\vtheta$ after observing the data~$\mathcal{D}$. Utilizing the posterior distribution $p(\vtheta \mid \mathcal{D})$, one can estimate a distribution for the predictions as follows
\begin{align*}
    p(\vy^* \mid \vx^*, \mathcal{D}) &= \int p(\vy^* \mid \vx^*, \vtheta) \, p(\vtheta \mid \mathcal{D}) \, \rm d \vtheta.
\end{align*}

Thus, the Bayesian approach allows for a more thorough representation of the prediction uncertainty, since it incorporates the uncertainty in estimating $\vtheta$. However, training a Bayesian Neural Network is a more complex process compared to training a standard neural network. This complexity arises from the fact that the posterior distribution $p(\vtheta \mid \mathcal{D})$ cannot be directly computed and must be approximated. 
Dropout as an approximation is one of the most widely adopted approaches for BayNN. 

\subsection{Normalization Methods in Deep Learning}
Normalization techniques are crucial for stabilizing deep neural network training by making the feature distribution zero mean and unit variance. Specifically, for a 4D input tensor of shape $[N, C, H, W]$ common in computer vision, these methods differ in the dimensions that they normalize. Batch Normalization (BatchNorm) normalizes each feature across the batch dimension $N$ and spatial dimensions $H, W$, and can be represented as
\begin{equation}\label{eq:normalization}
    \hat{y} = \frac{y - \mu}{\sqrt{\sigma^2 + \epsilon}}.
\end{equation}
Here, $\mu$ and $\sigma$ represent the mean and standard deviation of the batch and $\epsilon$ is a small constant added for numerical stability.
Afterward, an affine transformation is performed on normalized activations, $\hat{y}\gamma+\beta$.
where $\gamma$ and $\beta$ are learnable parameters, referred to as affine parameters in the following. Layer Normalization (LayerNorm) normalizes each feature across the channel $C$ and the spatial dimensions $H, W$ for each instance in the batch. Instance Normalization (InstanceNorm) is tailored for style-transfer tasks and normalizes each channel for each data instance. Group Normalization (GroupNorm) divides the channels into groups and normalizes within each group. These methods are essential for different applications and hardware constraints.

\subsection{Dropout}
The conventional Dropout was designed as a regularization technique to prevent NN overfitting. During training, the Dropout randomly sets the input units to zero at each forward pass with a probability of $p$. This prevents the network from relying excessively on any single neuron. Specifically, for each neuron, Dropout generates a random binary variable drawn with probability $1-p$ from a Bernoulli distribution, also known as the Dropout mask. The output of a Dropout layer is calculated by multiplying the Dropout mask by the input.

There have been other variants of Dropout with different goals and capabilities. DropConnect, for instance, modifies the Dropout technique to operate on the weights. The Gaussian Dropout variant replaces dropped neurons with Gaussian noise, which can be interpreted as adding a measure of uncertainty to the Dropout procedure. On the other hand, spatial Dropout removes entire feature maps from the convolutional layers, making the network resilient to the loss of spatially correlated features.

Although these methods are effective in their respective domains, they offer limited inherent robustness to NVM non-idealities.
Our work addresses this deficiency by introducing \emph{Affine Dropout}, an innovative extension of conventional Dropout techniques.

\subsection{In-Memory Computation using NVMs}

IMC can efficiently be realized by the use of NVM technologies. In IMC, NVM cells are organized in a crossbar matrix. It allows the execution of weighted sum operations in an analogue manner directly within memory with $O(1)$ time complexity and eliminates the need for data transfers between processing elements and memory. 
Several NVMs have been used for such an approach, such as resistive
memory, magnetic memory, phase change memories, and flash memory \cite{wang2020resistive}. 
However, NVM devices exhibit non-ideal characteristics such as conductance variation, retention errors, and quantization errors. These challenges become apparent when attempting to implement traditional NNs with their default 32-bit floating-point precision parameters in crossbar arrays, as NVMs have limited stable conductance states. To adapt these parameters for efficient deployment in IMC and mitigate performance losses, they are quantized during training to lower bit resolutions.

In this process, each bit of the pre-tranined quantized NN parameters is mapped to a conductance value in NVM cells within the crossbar array before deployment. Input vectors are converted into continuous voltages and fed into the array, where the resulting current represents the weighted sum. This sum is then digitized, with data conversion circuits like Digital-to-Analog Converters (DACs) and Analog-to-Digital Converters (ADCs) facilitating these steps. Subsequently, additional computational operations such as bias addition and batch normalization are performed on digital values.

\subsection{Related Works}
In the context of the Bayesian framework, the work~\cite{9586160} modeled the variance of RRAM and used this as the prior distribution for training. Their approach needs variation modeling followed by training for every NVM technology and may not work for other non-idealities such as programming errors. Also, the uncertainty estimation is ignored during inference.

The work in~\cite{9586115} proposed a Bayesian optimization method with a NAS to find a robust network for weight drift non-ideality. They showed that the conventional and alpha Dropout layer provides better robustness to weight drift compared to various normalization and activation functions. Subsequently, they search for the Dropout rate, which gives better robustness. However, their evaluation shows lower baseline accuracy on harder tasks or larger models. Also, a NAS, which is itself highly computation-heavy, may need to be carried out for each type of non-idealities and NVM technology. Similar work~\cite{ye2023improving} proposed a noise injection approach, involving the placement of noise injection layer after each layer, excluding the final softmax layer. They used NAS to find a proper noise injection setting and explore different types of noise, such as Gaussian, Laplacian, and Dropout. However, their evaluation shows severe baseline accuracy degradation with Laplacian noise injection. In addition, their method has similar drawbacks as~\cite{9586115}.

In conventional NNs, several works aim to improve the robustness of IMC. In one group of work, the non-idealities of the IMC are modeled and injected during training~\cite{9586115}. Although this can improve robustness for specific nonidealities that were injected, it may not translate into other non-idealities such as manufacturing defects or retention faults. Also, it requires modeling for each NVM technology or hardware device which is not generalizable. 

Other works propose to retrain or recalibrate the NN to improve robustness~\cite{8988622, tsai2020robust}. However, those approaches have a high overhead as retraining and recalibration can be computationally demanding, and recalibration data need to be stored in hardware, which incurs extra memory. Although works such as \cite{10103064} propose to reduce those costs, it still needs to be done periodically for each device.


\section{Methodology}\label{sec:methodology}
\begin{figure}
    \centering
    \includegraphics[width=0.9\linewidth]{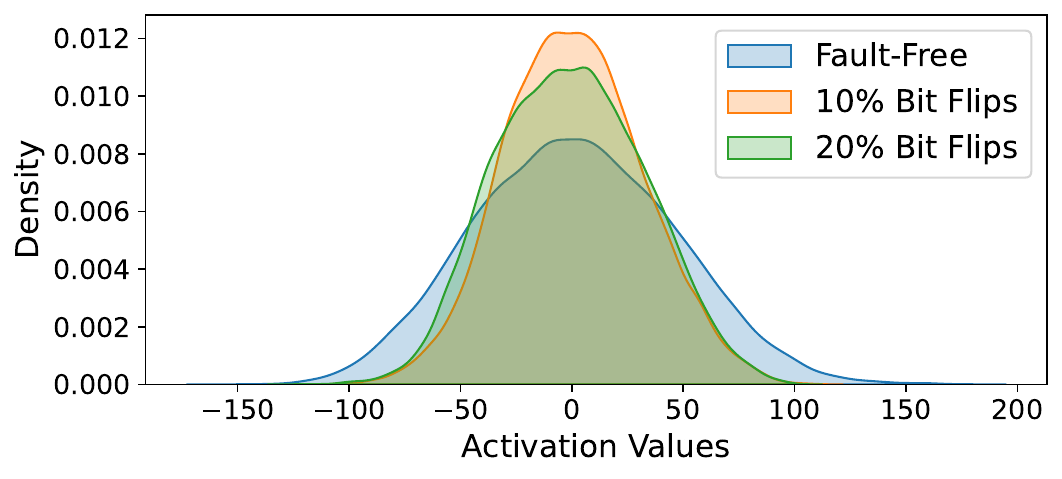}
    \caption{Change in activation distribution due to faults.}
    \label{fig:KDE_Bit_flips}
\end{figure}

In existing work~\cite{tsai2020robust, joshi2020accurate}, it has been shown that due to the non-idealities of the NVMs, the distribution of weighted sum shifts from the trained distribution. 
Figure~\ref{fig:KDE_Bit_flips} shows the distribution change due to $10\%$ and $20\%$ bit-flip faults. 
Based on this observation, it can be stated that \emph{non-idealities of NVMs} adds additive or multiplicative noise to the weighted sum of a layer. Existing work~\cite{kim2020efficient} also supported this statement based on their NVM variations model.

Therefore, we hypothesize that adding stochastic additive and multiplicative components to the weighted sum of a layer would increase the robustness of an NN against these types of noise. To this end, we introduce \emph{inverted normalization and affine Dropout}. In our approach, the order of the affine transformation in the normalization layer is reversed, with the affine parameters randomly dropped. Furthermore, as mentioned in~\cite{ye2023improving}, the scaling factors ($\gamma$) of the normalization layers amplify the drift of the parameters and, in turn, reduce the accuracy of the network. Thus, treating them as random parameters can potentially improve robustness.

In addition, in our approach, normalization is done for each output instance or group of neurons in a layer. This adds another layer of robustness by standardizing the weighted sum of each layer in the case of distribution shifts (see Fig.~\ref{fig:KDE_Bit_flips}) due to non-idealities. This approach has been proven to be effective in improving robustness in works~\cite{joshi2020accurate}. 

Compared to the conventional normalization method, we treat the normalization layer differently. As mentioned previously, our primary focus is on enhancing the robustness and accuracy of BayNNs without reducing their accuracy and uncertainty estimation capabilities. Various components of our method are described in detail in the following sections.




\subsection{Inverted Normalization}

As stated in Section~\ref{sec:methodology}, normalization techniques such as batch normalization or layer normalization adjust the input to zero mean and unit variance in different dimensions. Subsequently, an \emph{optional} affine transformation is carried out using learnable parameters. The main aim was to give the NN freedom to reverse the normalization if it is beneficial. However, in practice, it is unlikely that the affine parameters will learn to reverse the normalization rather than participate in the optimization process. 
We have observed the distribution before and after the affine transformation of the normalization layers of several topologies and found that they are different.

Based on this observation, we propose the inverted normalization layer. In our approach, we treat the affine parameters of the normalization as parameters similar to the weights and biases of the NN. That is, their learning objective only is to minimize the loss using the gradient descent algorithm. For simplicity, in the following, the affine parameters $\gamma$ and $\beta$ are referred to as weights and biases of the inverted normalization layer.

\begin{figure}
    \centering
    \includegraphics[width=\linewidth]{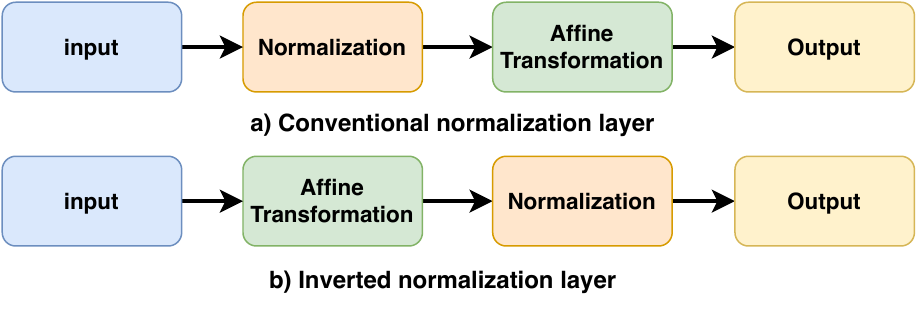}
    \caption{Computation flow of the proposed and conventional normalization layers.}
    \label{fig:inverterNorm_flow}
\end{figure}

Also, unlike traditional normalization methods, the affine transformation in our approach is \emph{necessary} and is done before normalization. Since normalization is still performed on the transformed input, the learning process remains stable. The supporting results are shown in Section~\ref{sec:results}.
Figure~\ref{fig:inverterNorm_flow} shows the flow diagram for the conventional and proposed inverted normalization layers. 
\setlength\belowcaptionskip{-5pt}

\begin{figure}
    \centering
    \includegraphics[width=\linewidth]{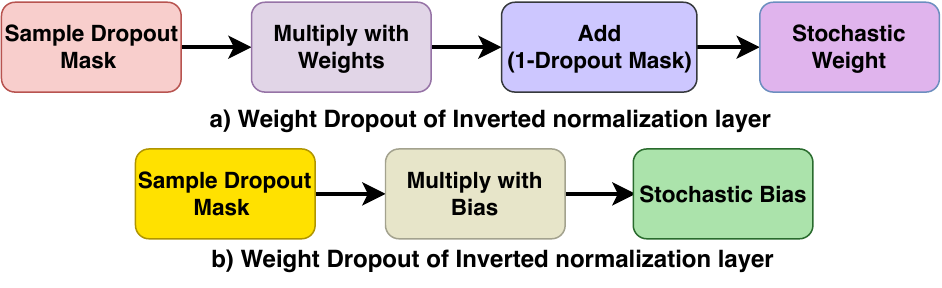}
    \caption{Operation flow for the proposed affine parameters (weight and biases) Dropout.}
    \label{fig:Affine_drop_flow}
\end{figure}


\subsection{Affine Dropout}

To add stochasticity to the weighted sum, we randomly drop the weight and bias of the inverted normalization layer independently with probability $p$. Unlike traditional Dropout techniques, the weight and bias of the inverted layer normalization are dropped to \emph{ones} and \emph{zeros}, respectively. Since the weights of the inverted normalization layer multiply the weighted sum, it cannot be dropped to zero. Dropping the weights to ones and zeros effectively ignores the weights and biases. 

To implement the proposed affine Dropout, two binary ($[0, 1]$) Dropout masks are sampled from the Bernoulli distribution with Dropout probability $p$. Subsequently, the masks are multiplied by the weights and biases of the inverted normalization layer. This will set dropped weights and biases to zeros. Lastly, a (1-Dropout mask) is added to the masked weights to ensure that the dropped weights are one. The flow diagram of the proposed affine Dropout is shown in Fig.~\ref{fig:Affine_drop_flow}. Subsequently, the affine transformation and normalization are performed as shown in Fig.~\ref{fig:inverterNorm_flow} (a).

Individual components of the weights and biases can be dropped independently, referred to as the element-wise Dropout. Alternatively, the entire weight and bias vector can be dropped at the same time, referred to as the vector-wise Dropout. Vector-wise Dropout is more efficient as it only needs to sample one Dropout mask for the entire weights or bias vector, irrespective of their length. In addition, it only needs \emph{one} random number generator per layer to implement Dropout in the IMC architecture. Therefore, in this paper, we employ vector-wise Dropout.

\subsection{Initialization of Affine Parameters}
Initialization of weights and biases is important to achieve not only comparable accuracy but also to allow the proper learning of the affine parameters. Traditional normalization methods initialize $\gamma$ and $\beta$ as ones and zeros, respectively. However, our inverted normalization initializes its weight and biases randomly. Otherwise, the initial weights and biases can produce the same gradient and update in the same way throughout the training. Also, random initialization allows for more randomness in the weighted sum, which can potentially improve robustness.

Specifically, the weights are initialized from a normal distribution with a mean and variance of $1$ and $\sigma_\gamma$, $\mathcal{N}(1, \sigma_\gamma)$. Initially, this either scales up or down the weighted sum randomly. Similarly, biases are sampled from a normal distribution $\mathcal{N}(0, \sigma_\beta)$. Alternatively, weights and biases can be initialized from uniform distributions $\mathcal{U}(0, k_\gamma)$ and $\mathcal{U}(-K_\beta, K_\beta)$, respectively.

\subsection{Bayesian Inference and Uncertainty Estimation}

In the Bayesian paradigm, every weight in the network is modeled as a probability distribution, which allows us to capture model uncertainty. As shown by Gal. et al.~\cite{gal2016Dropout}, a NN trained with conventional Dropout and weight decay (L2 regularization) is an approximation of a Gaussian process. During inference, multiple forward passes through the network, each time sampling different Dropout masks, will give a stochastic output distribution. The average of all of these outputs gives the final prediction.

Our proposed affine Dropout acts as an alternative to the conventional Dropout in a Bayesian setting. Multiple forward passes can be made through the network with each time independently sampling weight and bias masks for each layer will give an output distribution. The final prediction is obtained from the average of those outputs. Uncertainty in prediction can be obtained from the variance of outputs or by calculating negative log-likelihood (NLL) for a classification task.

By integrating Bayesian inference with our inverted normalization and affine Dropout techniques, we present a model that is not only robust to various forms of noise and non-idealities but is also capable of quantifying the uncertainty associated with its predictions. This makes it particularly suitable for deployment in safety-critical applications where both performance and reliability are crucial.

\section{Results}\label{sec:results}
\subsection{Simulation Setup}
\subsubsection{Evaluated Models and Training Settings}
To show generalizability, we have evaluated our method on four different deep learning tasks: image classification, audio classification, autoregressive time series forecast, and semantic segmentation. For image classification, we used the CIFAR-10 benchmark dataset on the ResNet-18 2D-convolutional NN (CNN) topology. The Google speech command dataset is evaluated on a five-layer 1D-CNN model, referred to as M5, for the audio classification task. Also, an NN with two LSTM layers and a classifier layer was used for the time-series forecast. Lastly, the DRIVE (digital retinal images for vessel extraction) dataset is trained on the popular U-Net topology for biomedical semantic segmentation tasks.

In terms of bit-precision, ResNet-18 is binarized using the algorithm~\cite{qin2020forward}. Semantic segmentation is a much harder task, thus activations of the U-net model are quantized to 4 bits using~\cite{choi2018pact}, but their weights are binarized. On the other hand, the LSTM and M5 models are quantized to 8 bits to show the generalizability of our approach in a range of bit precision. Detailed training settings will be published in the GitHub repository\footnote{https://t.ly/iA2rM}.

The proposed inverted normalization layer is applied following all the convolutional layers as a drop-in replacement for conventional normalization layers. A Dropout probability of $0.3$ is used for all our models. In U-Net, we have normalized across groups of $\frac{C_{out}}{8}$ channels together with the same train-time and test-time behavior as Group Normalization. Here, $C_{out}$ refers to the output channels. In other models, we have normalized each input instance, the same as the Layer Normalization method.

\subsubsection{NVM Non-idealities Model}
In this work, we have abstracted the circuit-level details into an algorithmic model in the Pytorch framework. We modeled both the manufacturing and thermal conductance variation as an additive and multiplication noise, as suggested by~\cite{kim2020efficient}. Additive variation is modeled as $\mathcal{N}(0, \sigma)$ and multiplicative variation is modeled as $\mathcal{N}(0, \sigma)$. As done by works~\cite{kim2020efficient, 9586115}, those noises are injected into weights for 8-bit weights, but in the case of binary NNs, they are injected into normalized activations before applying the \textit{Sign(.)} function.

Other post-manufacturing and infield non-idealities of NVMs, such as programming errors, and retention faults are modeled as bit-flips. For binary and quantized parameters, the random bits are flipped in each simulation run. We also conducted a different experiment on the LSTM model, where we injected (random) uniform noise with varying strength.

We perform 100 Monte Carlo fault simulation runs, simulating 100 chip instances, for each variation and bit-flip scenario, and report the mean and standard deviation of accuracy.

\begin{figure}
    \centering
    \includegraphics[width=1\columnwidth]{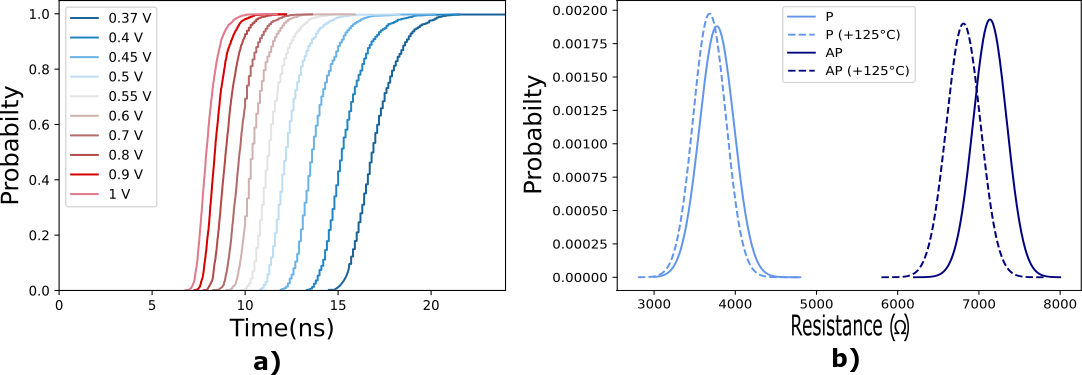}
    \caption{Examples of non-idealities: (a) Stochastic switching in magnetic memories under different voltages and (b) influence of temperature on the resistance distributions (Monte Carlo simulations).}
    \label{fig:switch_prob}
\end{figure}


\subsection{Baseline Inference Accuracy}
Although our aim is to improve robustness against the non-idealities of the NVMs, it is equally important to achieve an accuracy comparable to the baseline NN and SOTA Dropout-based bayNNs~\cite{spindrop, ahmed2023spatial}. As shown in Table~\ref{tab:acc}, the inference accuracy of our method is comparable to the baseline on all the datasets and topologies evaluated. In the worst case, the accuracy of CIFAR-10 is $0.66\%$ below the SpatialSpinDrop~\cite{ahmed2023spatial} method. However, using a smaller Dropout probability such as $0.1$ or $0.2$ can improve accuracy, but may be less robust to NVM non-idealities. However, the proposed approach outperforms the conventional NN in all metrics.

\setlength{\tabcolsep}{2pt}
\begin{table}[]
\footnotesize
\caption{Summary of inference accuracy of the proposed method and related works evaluated on different datasets, bit-precision, metrics, and topologies. Here, W/A refer to the bit-precision of weights and activation.}
\resizebox{\linewidth}{!}{%
\begin{tabular}{|c|c|c|c|c|c|c|c|}
\hline
Topology  & Dataset                                                           & metrics & W/A & NN      & SpinDrop & SpatialSpinDrop & \textbf{Proposed} \\ \hline
ResNet-18 & CIFAR-10                                                          & Accuracy $ \uparrow$     & 1/1 & 89.01\% & 89.82\%    & 90.5\%            & \textbf{89.82\%}  \\ \hline
M5        & \begin{tabular}[c]{@{}c@{}}Google Speech \\ Commands\end{tabular} & Accuracy $ \uparrow$     & 8/8 & 83.97\% & 84.83\%    &           -        & \textbf{85.28\%}  \\ \hline
U-Net     & DRIVE                                                             & mIoU $ \uparrow$        & 1/4 & 66.87\% & 67.93\%    & 64.6\%            & \textbf{67.54\%}  \\ \hline
LSTM      & Atmospheric CO2                                                   & RMSE $ \downarrow$        & 8/8 & 0.1264  & 0.1534     & -                 & \textbf{0.1219}   \\ \hline
\end{tabular}
}
\label{tab:acc}
\end{table}

\subsection{Analysis of Variation Robustness}
We only compare our work with related Dropout-based BayNNs targetting IMC hardware. Variational inference-based works~\cite{soyed_DATE23, ahmed2023spinbayes} represent completely different learning and network topologies. Therefore, they are ignored for comparison.

\begin{figure*}
\centering
\begin{subfigure}{.5\textwidth}
  \centering
  \includegraphics[width=\linewidth]{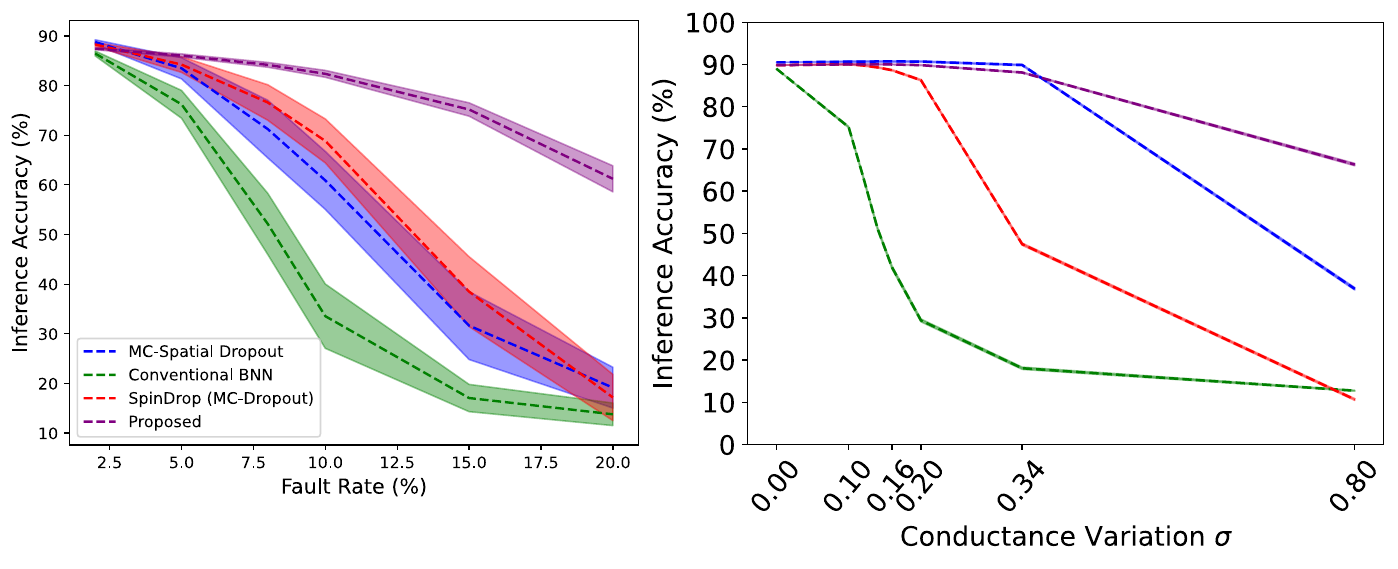}
  \caption{CIFAR-10 dataset on ResNet-18}
  \label{fig:CIFAR}
\end{subfigure}%
\begin{subfigure}{.5\textwidth}
  \centering
  \includegraphics[width=\linewidth]{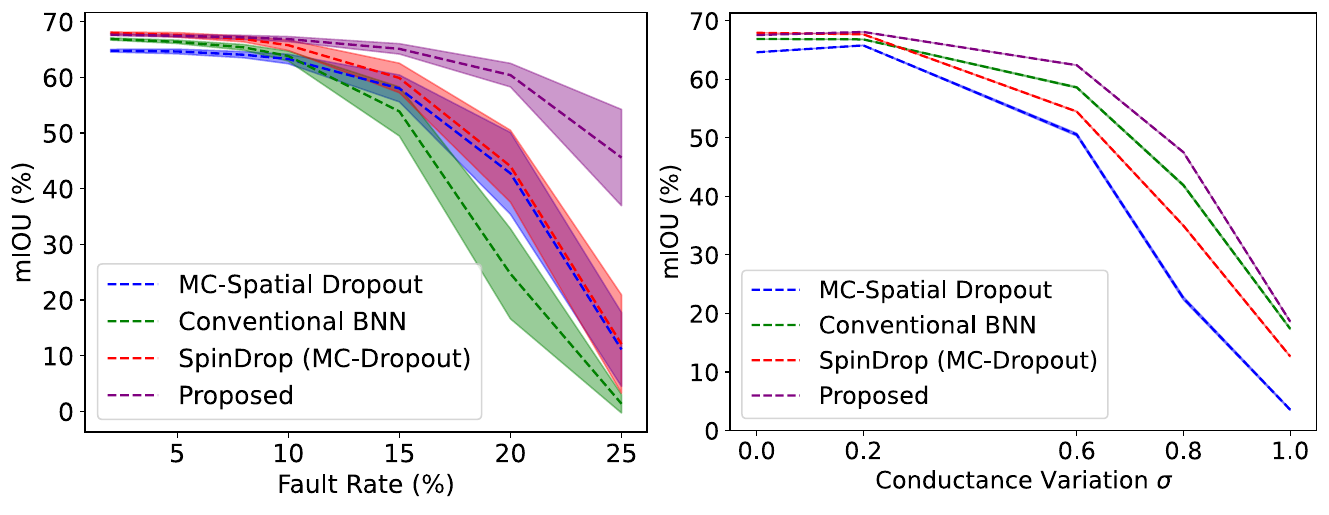}
  \caption{DRIVE dataset on U-Net}
  \label{fig:DRIVE}
\end{subfigure}
\footnotesize
\caption{Evaluation of robustness of ResNet-18 and U-Net topologies on CIFAR-10 and DRIVE datasets. The shaded region shows $\pm$ one standard deviation variation from the mean. The left and right figures of both datasets illustrate the evaluation of bit-flips and additive conductance variations, respectively.  }
\label{fig:CIFAR_DRIVE}
\vspace{-1em}
\end{figure*}

\begin{figure*}
\centering
\begin{subfigure}{.42\textwidth}
  \centering
  \includegraphics[width=\linewidth]{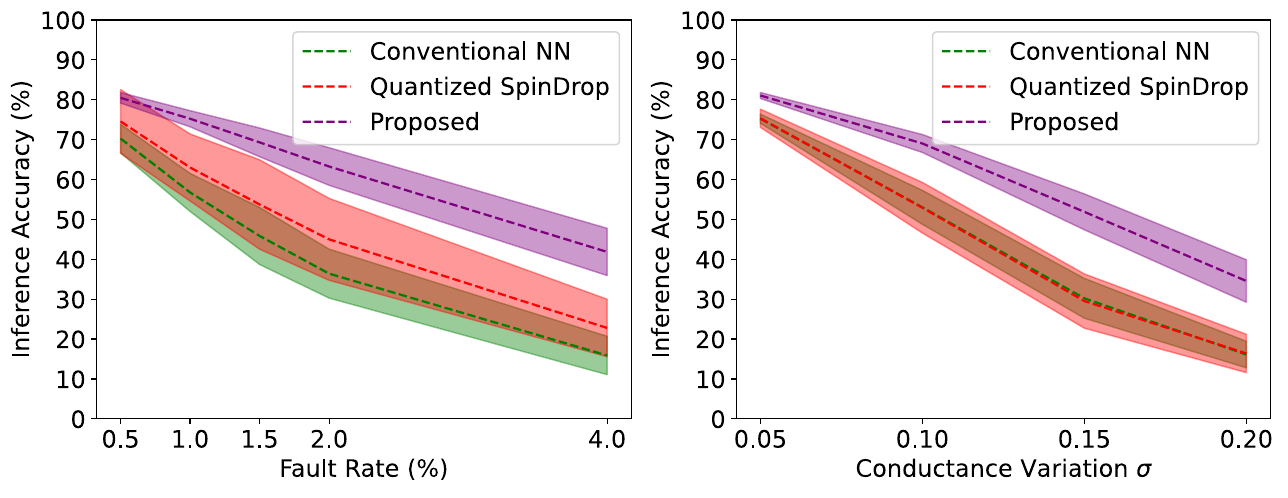}
  \caption{ Googe speech commands dataset on M5}
  \label{fig:DATE24_Speech}
\end{subfigure}%
\begin{subfigure}{.6\textwidth}
  \centering
  \includegraphics[width=\linewidth]{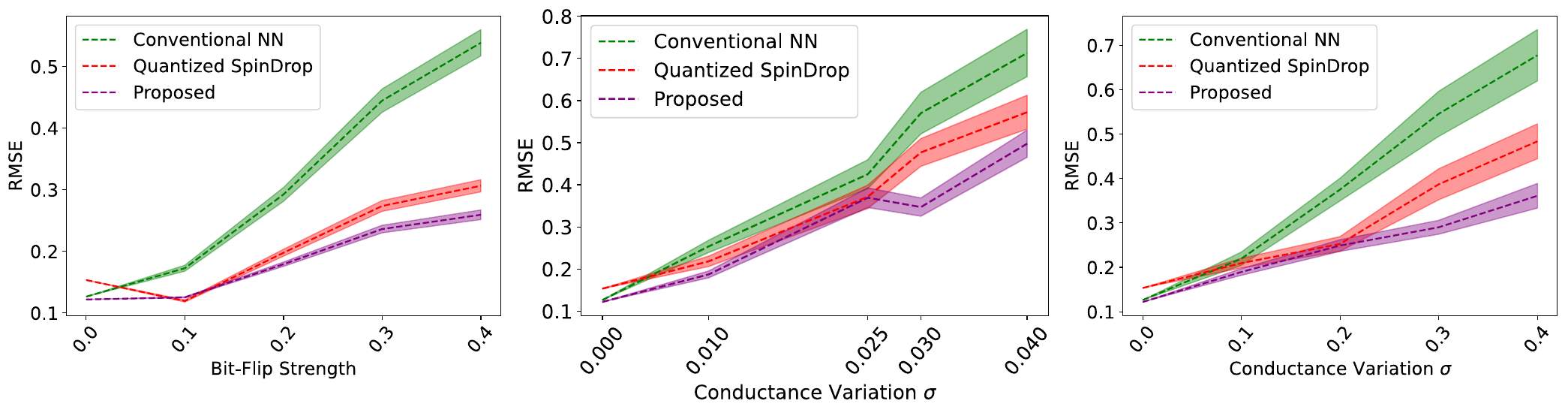}
  \caption{Atmospheric CO2 forecast with LSTM-based autoregressive model}
  \label{fig:DATE24_CO2}
\end{subfigure}
\caption{Evaluation of conductance and bit-flip robustness of ResNet-18 and U-Net topologies on CIFAR-10 and DRIVE datasets. The shaded region shows $\pm$ one standard deviation variation from the mean. Here, the first and second figures of both datasets show the evaluation of bit-flips and additive conductance variations, respectively. In (b), the last figure shows multiplicative conductance variations. }
\label{fig:CO2_Speetch}
\vspace{-1em}
\end{figure*}

In terms of robustness to NVM conductance variations, our proposed approach is robust to additive and multiplicative variations on all datasets with varying degrees of variation. As shown in Figs~\ref{fig:CIFAR_DRIVE} and ~\ref{fig:CO2_Speetch}, our approach leads to a graceful degradation in accuracy or other metrics evaluated. Specifically, our approach improves inference accuracy by up to 55.62\% compared to other Dropout-based BayNN approaches and 53.55\% compared to conventional NN. In the case of semantic segmentation, there is a marginal improvement in accuracy. In LSTM-based time series prediction, the RMSE score is reduced by up to 30.2\% and 46.7\% for additive and multiplicative variations, respectively.

We have performed a detailed evaluation of additive variations, but only in the LSTM model, an evaluation of multiplicative variations was performed. Nevertheless, our results translate to multiplicative variations on other datasets.

\subsection{Analysis of Bit-flip faults Robustness}
Similarly to conductance variations, our method shows significant robustness to bit-flip faults in all datasets. Our approach can improve accuracy by up to $42.06\%$ compared to other Dropout-based BayNN approaches and $58.11\%$ compared to conventional NN. Also, the standard deviation in accuracy is smaller for our approach compared to other approaches, as shown by the narrower band in Fig.~\ref{fig:CIFAR_DRIVE} and~\ref{fig:CO2_Speetch}. In the LSTM model, our approach reduces the RSME score by up to $51.84\%$.

\subsection{Uncertainty Estimation}
Typically, it is assumed that the training data and the test data are derived from the same distribution (ID). However, when these distributions are not aligned, such as when the test images are rotated or contaminated with measurement noise, the model's predictive uncertainty should increase to indicate that the model's prediction is dubious. These data are termed out-of-distribution (OOD) data.

\begin{figure}
    \centering
    \includegraphics[width=\linewidth]{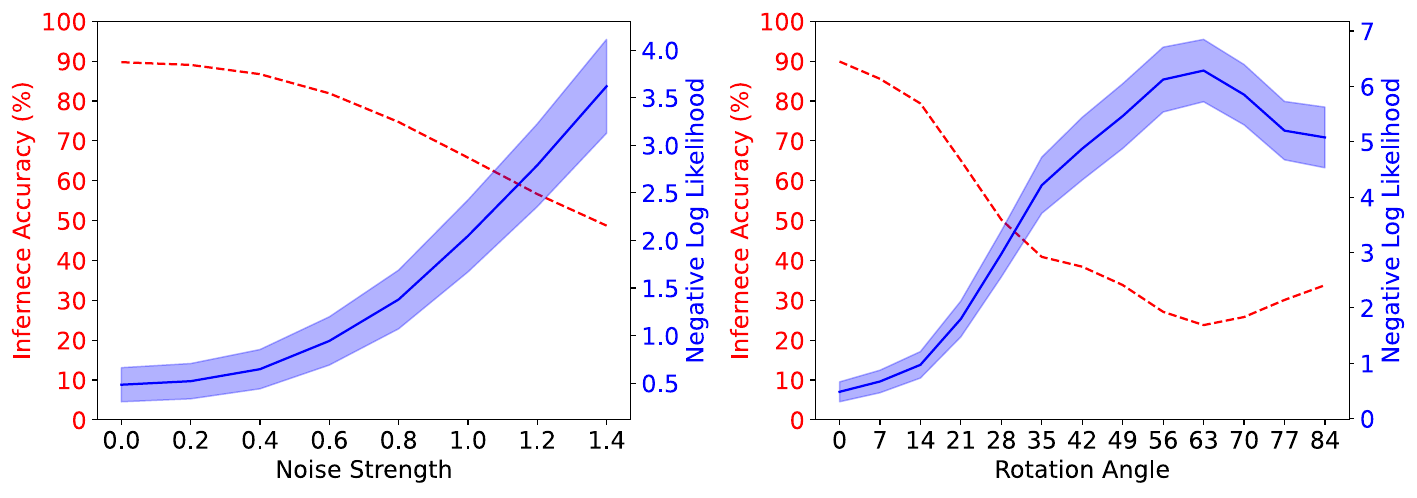}
    \caption{Evaluation of the proposed method on OOD data. (Left) Uniform noise is added to images and (right) images are rotated to shift the distribution.}
    \label{fig:DATE24_uncer}
    \vspace{.5em}
\end{figure}

To evaluate this feature of the proposed BayNN, we conducted two identical experiments to~\cite{soyed_DATE23} to investigate the effect of shifting the dataset in various ways. In the first experiment, images were gradually rotated in 7-degree increments in 12 stages. In the second scenario, escalating random uniform noise levels were introduced. As depicted in Fig.~\ref{fig:DATE24_uncer}, as a consequence of these shifts in the data distribution, the accuracy of the inference decreases and the NLL score increases. We use NLL as an uncertainty estimation metric, similar to work~\cite{soyed_DATE23}. In general, a lower NLL score is desired for ID data and a higher NLL score for OOD data.

The NLL score can be utilised to detect OOD data. A value greater than a predetermined threshold, such as the average NLL score on the test dataset, indicates the presence of OOD data. Using this methodology, we can identify up to $55.03\%$ and $78.95\%$ of OOD instances for uniform and random rotation experiments, an improvement of $14.61\%$ compared to~\cite{soyed_DATE23}.

\subsection{Impact of Initialization on Inference Accuracy}
Initialization of the weights and biases of the proposed inverted normalization is important to improve not only the accuracy of fault-free inference but also the robustness against NVM non-idealities. 

The weights and biases of the proposed inverted normalization are initialized from normal distributions with a value of $0.3$ for $\sigma_\gamma$ and $\sigma_\beta$. We have found that initializing with larger $\sigma_\gamma$ and $\sigma_\beta$ can improve robustness to variations and bit-flip faults, as it introduces more randomness to the weighted sum. However, it can reduce the accuracy of baseline by $1$-$2\%$. 


\section{Conclusion}\label{sec:conclusion}
In this paper, we present a self-immuned Bayesian neural network framework for reliable IMC implementation in safety-critical applications. We introduce an inverted normalization layer that performs the affine transformation first, then normalization. Also, we propose the affine dropout, which introduces randomness into the weighted sum of each layer where it is applied. In turn, the combined effect of those leads to inherent robustness to NVM non-idealities in their IMC implementation. 
We show that the fault-free prediction performance is comparable to that of SOTA BayNN and conventional NN with different parameter precisions and deep learning tasks. Additionally, in various tasks and non-ideality scenarios, our approach can improve inference accuracy by up to $58.11\%$ and RMSE by up to $51.84\%$. Nevertheless, our approach does not compromise the uncertainty estimation capabilities of BayNN, with up to $78.95\%$ detection of OOD instances.

\bibliographystyle{IEEEtran}
\bibliography{bibliography}

\end{document}